\documentclass[11pt, journal,onecolumn]{IEEEtran}
\IEEEoverridecommandlockouts
\usepackage{cite}
\usepackage{amsmath,amssymb,amsfonts}
\usepackage{booktabs}
\usepackage{graphicx}
\usepackage{textcomp}
\usepackage{xcolor}
\usepackage{graphicx}
\usepackage{amsmath, xspace,  comment}
\usepackage{tikz, xcolor}
\usepackage{algpseudocode}
\usepackage[linesnumbered]{algorithm2e}
\usepackage{colortbl}
\usepackage{textcase} 
\usepackage{enumitem}
\usepackage{balance}
\usepackage{graphicx}
\usepackage{multirow}
\usepackage{hhline}
\usepackage{caption}
\usepackage[caption=false]{subfig}
\usepackage{soul}
\usepackage{hyperref}
\usepackage{makecell}
\usepackage[a4paper,left=1.2in,top=1.2in,right=1in,bottom=1in,nohead]{geometry}

\def\BibTeX{{\rm B\kern-.05em{\sc i\kern-.025em b}\kern-.08em
    T\kern-.1667em\lower.7ex\hbox{E}\kern-.125emX}}

\let\svthefootnote\thefootnote
\newcommand\freefootnote[1]{%
\let\thefootnote\relax%
\footnotetext{#1}%
\let\thefootnote\svthefootnote%
}
\newcommand*\cib[1]{\tikz[baseline=(char.base)]{
                            \node[shape=circle,fill=black,text=white,draw,inner sep=0.3pt] (char) {#1};}}

\newcommand{\etal}{{\em et al.}\xspace}
\newcommand{\eg}{{\em e.g.,}\xspace}
\newcommand{\ie}{{\em i.e.,}\xspace}

\newcommand{\BfPara}[1]{{\vspace{1ex}\noindent{\bf \em#1.}}\xspace}
\usepackage{tcolorbox}
\tcbuselibrary{theorems}

\newtcbtheorem[]{observation}{\textbf{Observation}}%
{colback=white,colframe=white!10!black,fonttitle=\bfseries,coltitle=white}{}

\hypersetup{
	colorlinks=true,
	urlcolor=blue!70!black,
	linkcolor=red!70!black,
	citecolor=purple,
} 

\begin{document}

\title{Impact of Architectural Modifications on Deep Learning Adversarial Robustness}

\author{
\IEEEauthorblockN{Firuz Juraev$^{1}$, Mohammed Abuhamad$^{2}$, Simon S. Woo$^{1,3}$,\\ George K Thiruvathukal$^{2}$, and Tamer Abuhmed$^{1}$\\}
\IEEEauthorblockA{$^{1}$ Department of Computer Science and Engineering, Sungkyunkwan University\\
$^{2}$ Department of Computer Science, Loyola University Chicago\\
$^{3}$ Department of Applied Data Science, Sungkyunkwan University\\
Emails: fjuraev@g.skku.edu, mabuhamad@luc.edu, swoo@g.skku.edu,\\ gkt@cs.luc.edu, tamer@skku.edu   
}}

\maketitle

\begin{abstract}
Rapid advancements of deep learning are accelerating adoption in a wide variety of applications, including safety-critical applications such as self-driving vehicles, drones, robots, and surveillance systems.
These advancements include applying variations of sophisticated techniques that improve the performance of models.
However, such models are not immune to adversarial manipulations, which can cause the system to misbehave and remain unnoticed by experts.
The frequency of modifications to existing deep learning models necessitates thorough analysis to determine the impact on models' robustness. 
In this work, we present an experimental evaluation of the effects of model modifications on deep learning model robustness using adversarial attacks.
Our methodology involves examining the robustness of variations of models against various adversarial attacks.
By conducting our experiments, we aim to shed light on the critical issue of maintaining the reliability and safety of deep learning models in safety- and security-critical applications.
Our results indicate the pressing demand for an in-depth assessment of the effects of model changes on the robustness of models\freefootnote{Corresponding author: Tamer Abuhmed (tamer@skku.edu)}.
\end{abstract}


\thispagestyle{plain}
\pagenumbering{gobble}
\section{Introduction}
Deep Learning (DL) has proven to be exceptional at solving real-world problems that traditional Machine Learning (ML) approaches could not address effectively \cite{abuhamad2021large,ali2022effective}. The DL methods have achieved significant advances in the traditional domains of computer vision, especially with the emergence of Deep Neural Networks (DNNs) and the availability of high-performance hardware to train complicated models \cite{chakraborty2021survey}. Important issues in the DL methods have also been disclosed simultaneously. One example is its vulnerability to undetectable input perturbations at test time \cite{abdukhamidov2023hardening, juraev2022depth}. 
For the image classification task, Szegedy \etal \cite{szegedy2014intriguing} initially developed minor perturbations on the input image, which tricked state-of-the-art DNNs with a high confidence score. The perturbed images are commonly-known as adversarial samples. This has led to a new dimension to the adversarial DL paradigm, which entails creating DNNs that are robust to such adversarial samples. Several factors must be addressed when designing robust networks, including what \textbf{\em perturbations} the adversary may apply to the input and what \textbf{\em knowledge} the adversary has about the model and the system. 

There are two types of adversarial attacks: \cib{1} training stage attacks and \cib{2} testing stage attacks. The \textit{training stage attacks} modify the training dataset \cite{barreno2006can}, the input features, or/and the data labels/classes in order to attack the target model. Any of these types of modification meet the definition of \textbf{perturbations} that must be understood in order for DNNs to be robust. The \textit{testing stage attacks} are two main types: \cib{A} white-box attacks and \cib{B} black-box attacks. The \textbf{white-box attack} is fully aware of the DL method, including its underlying structure and parameters. On the other hand, the \textbf{black-box attacks} cannot access knowledge about the target model, but they may train a local substitute model by querying the target model \cite{papernot2017practical}, exploiting the transferability of adversarial samples, or employing a model inversion approach.

The robustness of deep neural network models has been the subject of extensive research, with numerous studies investigating this issue from various angles \cite{juraev2022depth}. However, with the emergence of new versions and updates of existing models designed to improve performance, it is crucial to understand the impact of these changes on model robustness. For instance, the Inception model was improved by incorporating residual connections, resulting in state-of-the-art performance in the 2015 ILSVRC challenge. 
Many other advancements in the field have led to novel techniques that can be applied to various models to improve their performance.
In this study, we aim to investigate how variations of models affect their robustness against adversarial attacks. Specifically, we use three well-known white-box attacks (Fast Gradient Sign Method (FGSM) \cite{goodfellow2015explaining}, Projected Gradient Descent (PGD) \cite{kurakin2017adversarial}, and  Carlini \& Wagner (C\&W) \cite{carlini2017towards}) to evaluate the robustness of different versions of the VGG, Inception, and MobileNet architectures. Our main research question is \textit{``What are the effects of architectural modifications on models' robustness?''}     

\begin{figure*}[h]
    \centering
    \includegraphics[width=0.9\textwidth]{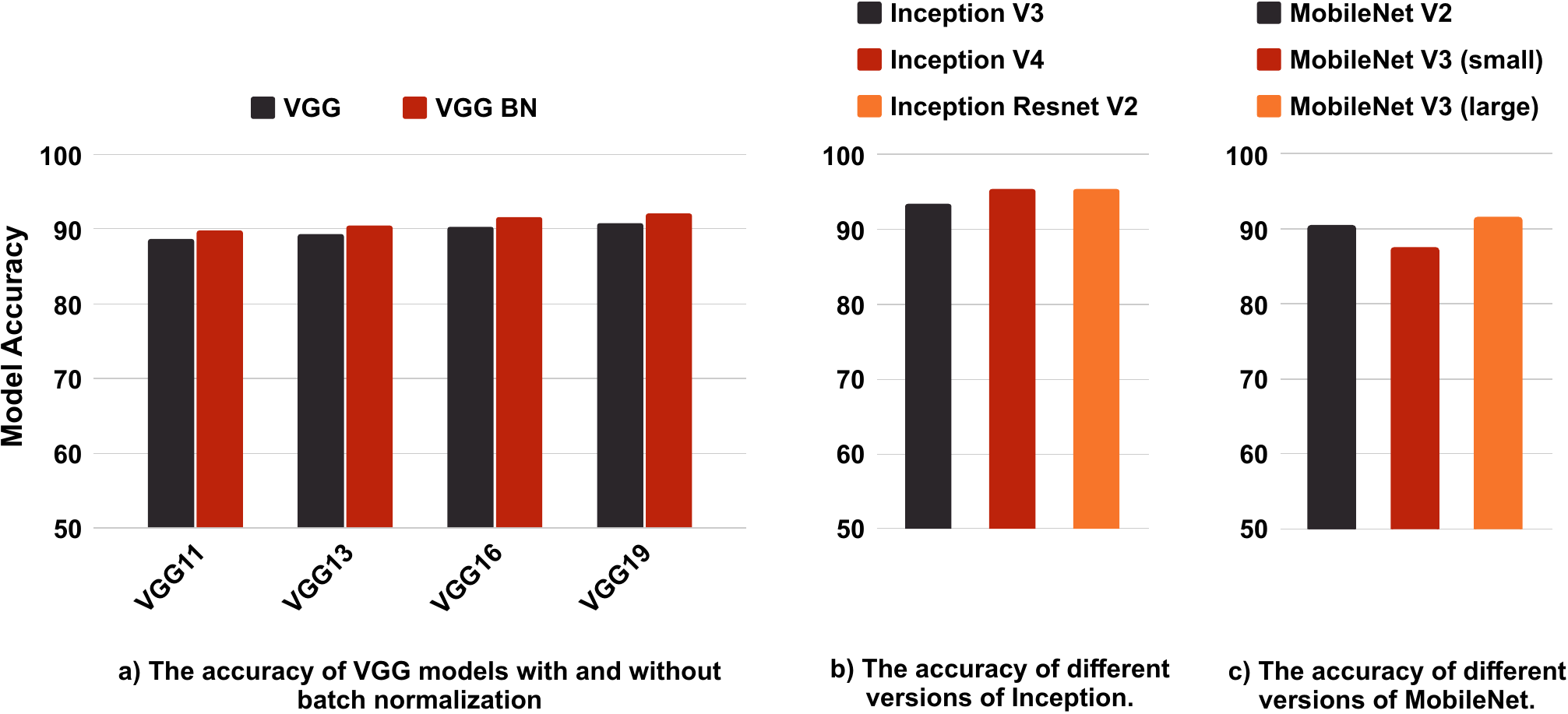}
    \caption{The accuracy of adopted models.}
    \label{fig:accuracy} 
\end{figure*}

\BfPara{Contributions} Our contributions are summarized as follows:
\begin{itemize}[label={}, leftmargin=2ex] 
    \item \cib{1}~ We investigate the impact of batch normalization on the performance and robustness of VGG-based models under adversarial scenarios. Through a series of experiments targeting variations of VGG models via various attacks, we highlight the necessity for careful consideration when using batch normalization, as its benefits in stabilizing and accelerating the training and improving the performance might also be associated with potential drawbacks in terms of robustness.
    \item \cib{2}~ We analyze variations of the Inception architectures, such as Inception V4 and Inception ResNet V2 under various adversarial scenarios. Our analysis shows that newer models, \eg Inception ResNet V2 exhibit improved performance and higher robustness to adversarial attacks.
    In particular, we show that these models can withstand adversarial attacks with longer attack times and higher noise rates, indicating a higher level of robustness. 
    \item \cib{3}~ We also explored small and large variants of MobileNet against adversarial scenarios. Our results suggest that MobileNet V3 outperforms MobileNet V2 while maintaining a high level of robustness. This can be contributed to the incorporation of the squeeze-and-excitation module (which allows the model to suppress noisy features) and hard-swish activation functions.
\end{itemize}

\BfPara{Organization}. The study is organized as follows. In Section \ref{sec:experiment_scenarios}, we discuss the dataset and studied models, Section \ref{sec:experiment_results} shows our results and observations, and in Section \ref{sec:discussion} and \ref{sec:conclusion} we provide some discussion and conclusion for our study.

\section{Experimental Settings} \label{sec:experiment_scenarios}
In this section, we discuss the settings of our experiments including the used dataset, models, and evaluation metrics.

\begin{table}[h]
\centering
\caption{The parameters of selected models.}
\label{tab:model_info}
\resizebox{0.65\textwidth}{!}{%
\begin{tabular}{l|ccc}
\toprule

\textbf{Model}       & \makecell{\textbf{Number of}\\ \textbf{parameters}} & \makecell{\textbf{Number of}\\ \textbf{Layers}} & \makecell{\textbf{Activation}\\ \textbf{functions}} \\ 
\midrule
VGG 11               & 132,863,336                   & 11                        & ReLU                          \\
VGG 13               & 133,047,848                   & 13                        & ReLU                          \\
VGG 16               & 138,357,544                   & 16                        & ReLU                          \\
VGG 19               & 143,667,240                   & 19                        & ReLU                          \\ \midrule
VGG 11 BN            & 132,868,840                   & 11                        & ReLU                          \\
VGG 13 BN            & 133,053,736                   & 13                        & ReLU                          \\
VGG 16 BN            & 138,365,992                   & 16                        & ReLU                          \\
VGG 19 BN            & 143,678,248                   & 19                        & ReLU                          \\ \midrule
Inception V3         & 27,161,264                    & 48                        & ReLU                          \\
Inception V4         & 42,679,816                    & 49                        & ReLU                          \\
Inception ResNet V2  & 55,843,464                    & 164                       & ReLU                          \\ \midrule
MobileNet V2         & 3,504,872                     & 53                        & ReLU                          \\
MobileNet V3 (small) & 2,542,856                     & 65                      & Hardswish                     \\
MobileNet V3 (large) & 5,483,032                     & 157                      & Hardswish                     \\ \bottomrule

\end{tabular}%
}
\end{table}

\BfPara{Dataset}  
In this work, we use the widely recognized and extensively used \textbf{ImageNet} dataset, which is composed of 14 million images. Each image in the dataset has a resolution of 224x224 pixels and belongs to one of 1000 classes.
For our experiments, we use 1000 test images for each attack. Those images are selected based on two conditions: \cib{1} one example from each class of ImageNet and \cib{2} each image must be classified correctly by all selected models.

\BfPara{Models} 
In this study, we employed a total of fourteen models from three commonly used families of convolutional neural networks: \textbf{VGG}, \textbf{Inception}, and \textbf{MobileNet}. Each family included different versions and types of models. We utilized pre-trained models on the PyTorch framework for our experiments on the ImageNet dataset.  
The characteristics and performance of the adopted pre-trained models are shown in Table \ref{tab:model_info} and Figure \ref{fig:accuracy}, respectively. 
In Table \ref{tab:model_info}, we show the key aspects of each model, such as the architecture, the number of layers, and the number of parameters. 
Figure 1 shows the top-5 accuracy of each model using ImageNet dataset.

\begin{figure*}[t]
    \centering
    \includegraphics[width=0.95\linewidth]{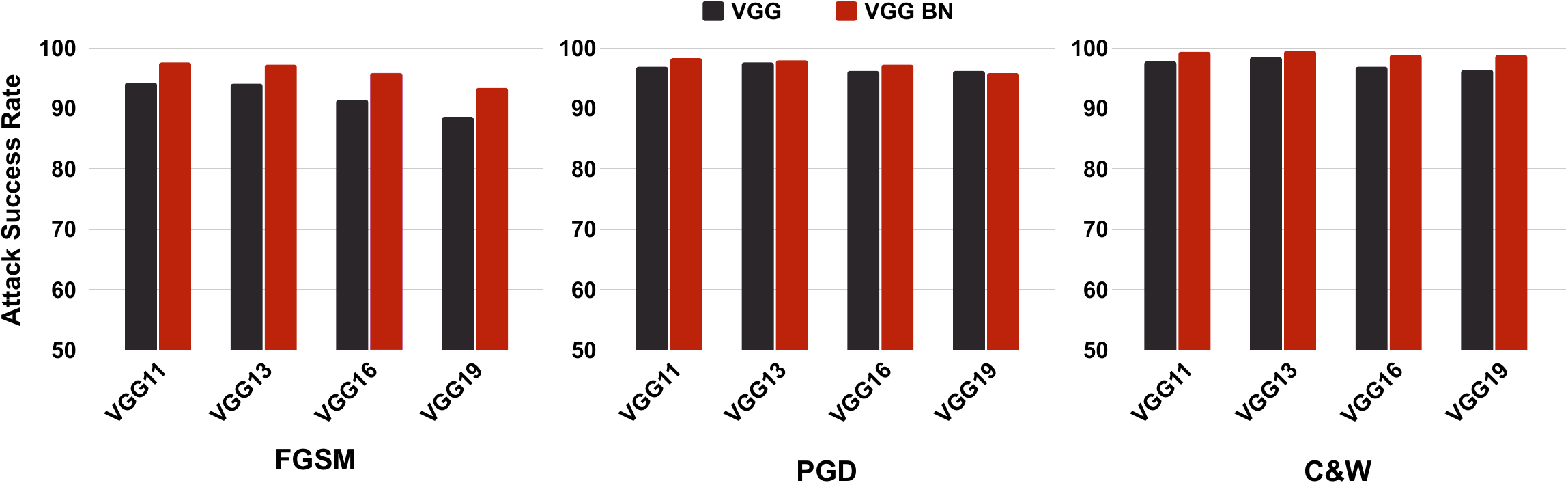}
    \caption{The impact of Batch normalization on VGG models.}
    \label{fig:vgg_bn} 
\end{figure*}
\begin{figure*}[t]
    \centering
    \includegraphics[width=0.95\linewidth]{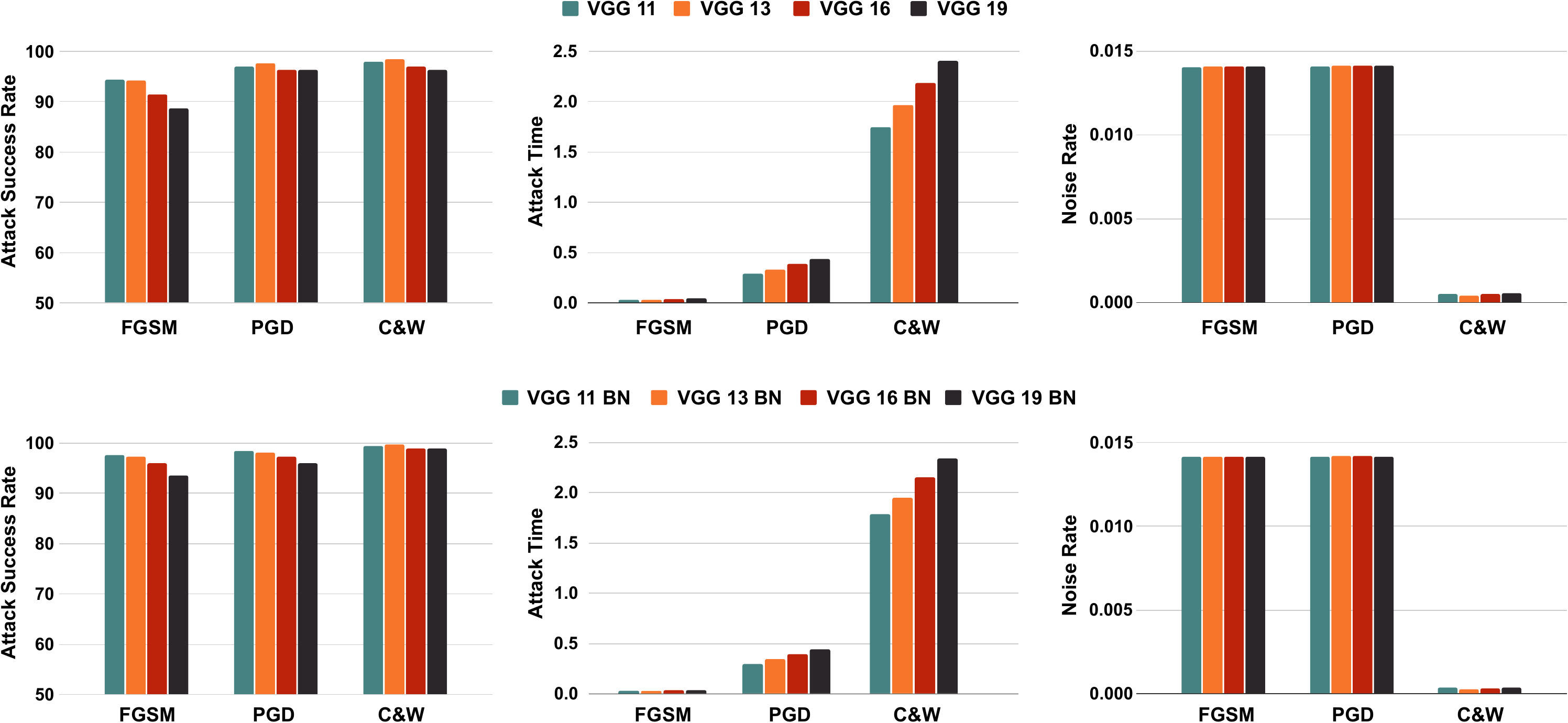}
    \caption{Attack success rate, attack time, and attack noise rate on VGG and VGG BN models.}
    \label{fig:vgg_three_metrics} 
\end{figure*} 

\BfPara{Evaluation Metrics}  
For evaluating the robustness of the models, we utilize the following evaluation metrics. 

\begin{itemize}[leftmargin=3ex]
    \item \textbf{Attack success rate:} This is calculated by dividing the number of successful attacks by the total number of attempts. 
    \item \textbf{Attack time:} We measured the time it took to add a specific level of noise to a benign image to achieve a successful attack. The results are presented as the average attack time in seconds for each model. 
    \item \textbf{Noise rate:} The amount of noise in the image is determined by using a metric called Structural Similarity Index Measure (SSIM) \cite{wang2004ssim}. SSIM measures how two images are similar to each other. For calculating the noise rate, we subtract the SSIM value from 1 (\ie noise rate = $1 - SSIM$). SSIM is calculated as follows.
    \begin{equation*}
    \label{eq:ssim}
    \text{SSIM}(x,y) = \frac{(2\mu_x\mu_y + c_1)(2\sigma_{xy} + c_2)}{(\mu_x^2+\mu_y^2+c_1)(\sigma_x^2+\sigma_y^2+c_2)}
    \end{equation*} 
    where  $x$ and $y$ are the two input images being compared, $\mu_x$ and $\mu_y$ are the mean values of $x$ and $y$, $\sigma_x$ and $\sigma_y$ are the standard deviations of $x$ and $y$, $\sigma_{xy}$ is the cross-covariance of $x$ and $y$, and $c_1$ and $c_2$ are small constants added to avoid division by zero.

\end{itemize}

\BfPara{Experiment Workstation} 
 The experiments are conducted on a machine equipped with an Intel Xeon(R) CPU E5-2620 v3 @ 2.40 GHz$\times$ 24 with Cuda-10.0 and three GEFORCE GTX TITANx 12 GB GPUs, as well as Python 3.7.7 distributed in Anaconda 4.8.3 (64-bit).

\begin{figure*}[ht]
    \centering
    \includegraphics[width=0.95\linewidth]{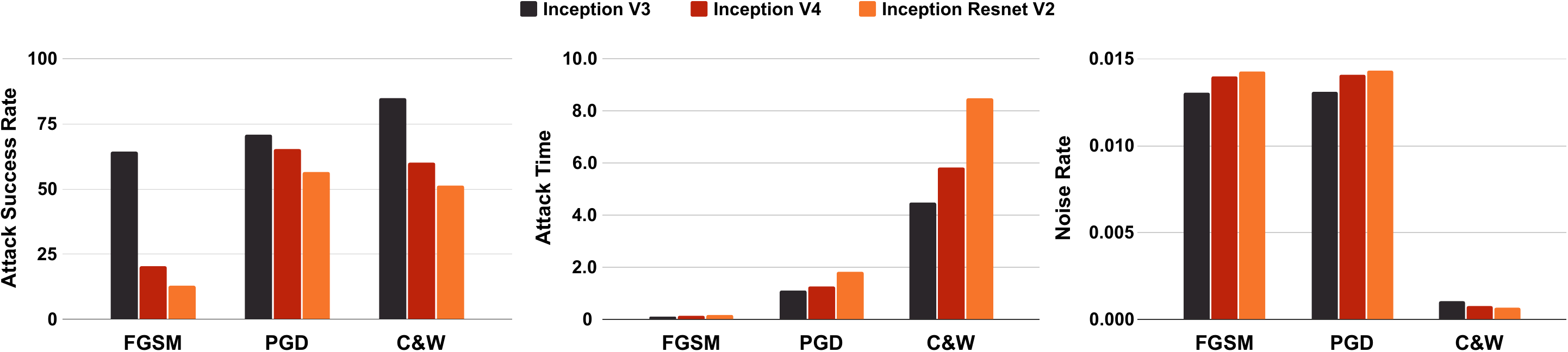}
    \caption{Attack success rate, attack time, and attack noise rate on Inception models.}
    \label{fig:inception} 
\end{figure*}

\begin{figure*}[ht]
    \centering
    \includegraphics[width=0.95\linewidth]{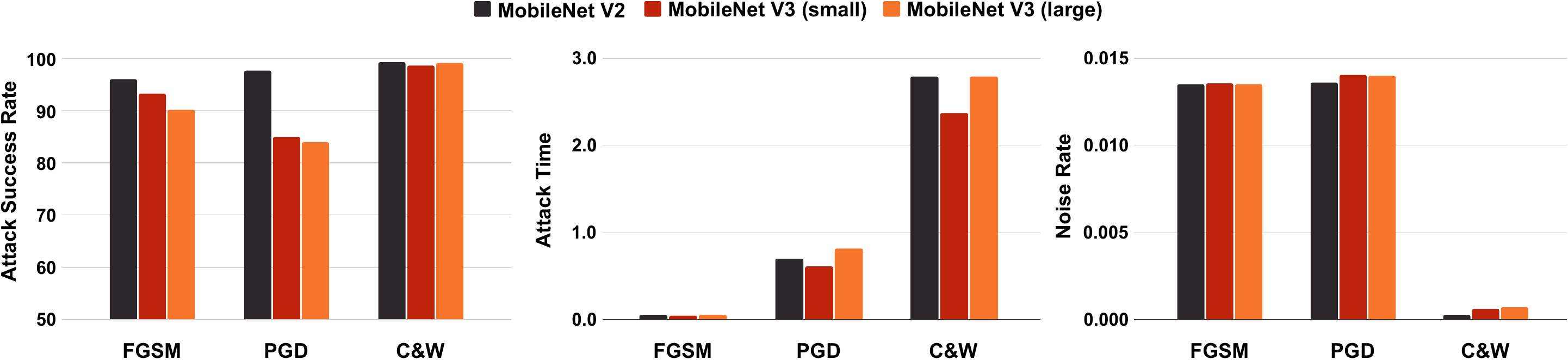}
    \caption{Attack success rate, attack time, and attack noise rate on MobileNet models.}
    \label{fig:mobilenet} 
\end{figure*} 

\section{Experimental results} 
\label{sec:experiment_results} 
In our experiments, we evaluated the robustness of the models using 1,000 images in terms of attack success rate, attack time, and noise rate. The main goal of the experiments is to examine the effects of architectural designs and techniques of deep neural models on performance and robustness. The study contributes to understanding the possible benefits and drawbacks of including various techniques in model designs, and how they impact the performance and robustness of the models. Particularly, we focus on analyzing the effects of batch normalization and architectural variations on three popular deep learning architectures, namely VGG, Inception, and MobileNet.

\subsection{Batch Normalization Effects on VGG models}

Batch normalization (BN) is a technique used in deep neural networks to increase training speed and stability. It works by normalizing each layer's inputs to have a zero mean and unit variance across a small batch of training data. BN is needed in deep neural networks to possibly address internal covariate shifts, improve stability, act as a form of regularization, and allow for higher learning rates.
Figure \ref{fig:accuracy}-(a) shows the comparison between the accuracy of VGG models with and without BN. The figure demonstrates that incorporating batch normalization improves the performance of VGG models. Specifically, the accuracy of VGG 11 increased from 88.75\% to 89.82\% when using BN. Similarly, in VGG 13 (increased from 89.26\% to 90.49\%), in VGG 16 (increased from 90.35\% to 91.61\%), and in VGG 19 (increased from 90.82\% to 92.07\%).

Although BN can enhance the stability and accuracy of the model and reduce overfitting, our results, as depicted in Figure \ref{fig:vgg_bn}, demonstrate that incorporating BN in VGG models can increase the success rate of adversarial attacks. 
In terms of attack time and noise rate, our results (as shown in Figure \ref{fig:vgg_three_metrics}) indicate that BN has an unnoticeable effect. 
One possible reason for this result is that BN can make the models more susceptible to gradient masking, which occurs when the gradients of the loss function with respect to the input become small or zero. This can contribute to decreasing the diversity of feature representations, which can make it difficult for the model to detect and respond to adversarial perturbations, leading to a decrease in overall robustness \cite{boenisch2021gradient}.

While BN can add additional parameters and non-linearity to a model, which may make it more difficult for adversarial attacks to succeed, our experiments have shown that VGG models with BN are actually more vulnerable. Specifically, when looking at FGSM attacks, we found that the attack success rate was 3.93 percentage points higher on VGG models with BN compared to those without BN. Similarly, on the PGD attack, VGG models with BN had a 0.63 percentage point higher attack success rate than models without BN. Interestingly, on the C\&W attack, we found that VGG models without BN were more robust than VGG models with BN, with the attack success rate being 1.80 percentage points lower on models without BN (Figure \ref{fig:vgg_bn}). 

Generally, BN in deep learning models can lead to improved performance and training speed, but our experiments suggest that it may negatively impact the model's robustness. Further exploration is necessary to better understand this trade-off, including investigating the impact of BN across a broader range of models, datasets, and attack scenarios.

In addition, we compared the performance of four different versions of the VGG model (\ie VGG 11, VGG 13, VGG 16, and VGG 19) to analyze how the number of layers in the model affects its robustness to adversarial attacks in Figure \ref{fig:vgg_three_metrics}. the results show that as the number of layers increases, the model becomes more robust as indicated by a decrease in attack success rate and an increase in attack time. However, the noise rate is stable across all versions.

\subsection{Architectural Updates Effects on Inceptions}  
The next experiment in this study was conducted using different versions of the Inception architecture, including Inception V3, Inception V4, and Inception ResNet V2. Both Inception V4 and Inception ResNet V2 demonstrated a superior performance of 95.3\% (top-5 accuracy) compared to Inception V3 with 93.45\% (Figure \ref{fig:accuracy}-(b), which was attributed to their various architectural improvements such as residual connections, more aggressive factorization, batch normalization, and bottleneck designs. Particularly, Inception V4 features a new ``stem'' module that reduces the computational cost \cite{szegedy2017inception}.

\begin{figure*}[ht]
    \centering
    \includegraphics[width=0.95\linewidth]{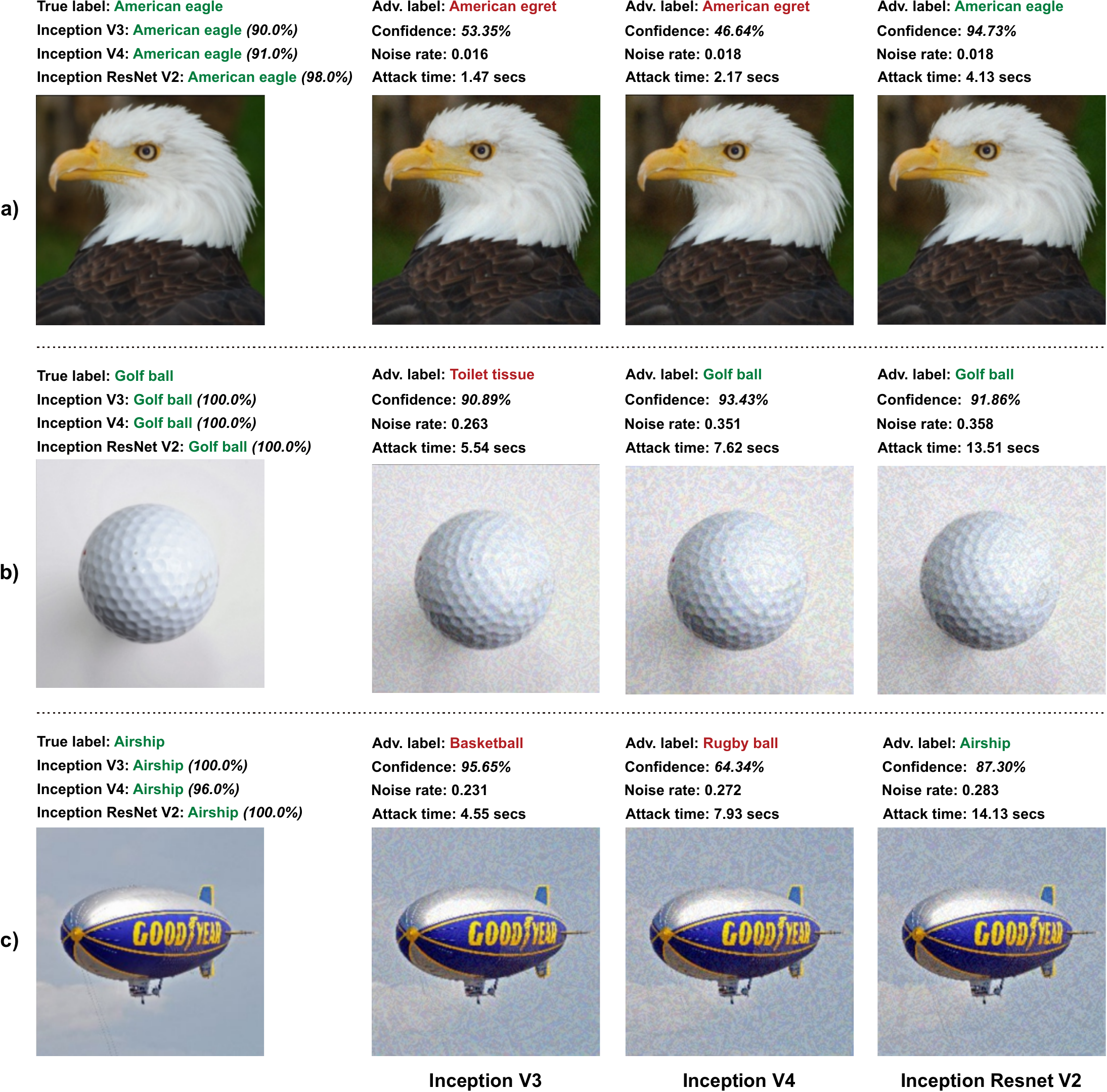}
    \caption{Amount of noise, confidence, and time needed by the PGD attack for Inception V3, Inception V4, and Inception ResNet V2 models. Images with a single object and a clear background ((B) and (C)) require more added noise and time.}
    \label{fig:example} 
\end{figure*} 

In Figure \ref{fig:inception}, Inception V4 and Inception ResNet V2 demonstrate higher robustness to adversarial attacks than Inception V3. The average attack success rates were 48.57\% and 40.23\% for Inception V4 and Inception ResNet V2, respectively, compared to 73.33\% for Inception V3. Inception V4 and Inception ResNet V2 also required longer attack times, with average times of 2.40 and 3.50 seconds, respectively, compared to 1.91 seconds for Inception V3. Additionally, Inception V4 and Inception ResNet V2 had higher average noise rates of 0.00961 and 0.00976, respectively, compared to 0.00907 for Inception V3, indicating their increased robustness to attacks.

To better understand the importance of additional metrics, such as attack time and noise rate, in evaluating model robustness, we provide examples in Figure \ref{fig:example}. In Figure \ref{fig:example}-(a), we show an example of a PGD attack on Inception V3, Inception V4, and Inception ResNet V2, demonstrating that as the model becomes more complex, it becomes more difficult to attack and requires more time to attack.
For instance, attacking Inception V3 took 1.47 seconds while attacking Inception V4 and Inception ResNet V2 took 2.17 and 4.13 seconds, respectively. 
However, the importance of noise in evaluating model robustness is difficult to distinguish in Figure \ref{fig:example}-(a) since it remains relatively constant across all models. Therefore, to better illustrate its importance, we allowed the attack to add more noise by increasing the epsilon (from 0.01 to 0.04) of the PGD attack and tested it on challenging images with a single object and a clear background, making it more difficult for the attack to succeed. 
Figure \ref{fig:example}-(b) and -(c) show that the attack adds more noise to the complex models, such as Inception V4 and Inception ResNet V2, compared to Inception V3.

\subsection{Architectural Updates Effects on MobileNets}  
In this experiment, we analyzed the performance of different versions of MobileNet, including MobileNet V2, MobileNet V3 (small), and MobileNet V3 (large). MobileNet V3 is an improved version of the MobileNet architecture, with both small and large variants available. MobileNet V3 demonstrated better performance compared to MobileNet V2 (Figure \ref{fig:accuracy} c), thanks to 
several enhancements made to the architecture \cite{howard2019searching}, including the integration of the squeeze-and-excitation module, the use of hard-swish activation functions, and improved batch normalization techniques. MobileNet V2 has a top-5 accuracy of 90.29\%, while MobileNet V3 (small) and (large) have an accuracy of 87.40\% and 91.34\%, respectively.

Our experiments, as shown in Figure \ref{fig:mobilenet}, show that MobileNet V3 (small) demonstrated superior robustness despite having significantly fewer parameters (2,542,856) than MobileNet V2 (3,504,872), and even though its accuracy is not higher than MobileNet V2. 
This suggests that MobileNet V3 (small) has more robust capabilities, making it more resilient against adversarial attacks. 
Specifically, MobileNet V2 had an average attack success rate of 97.63\% compared to MobileNet V3 (small) with 92.27\%. 
Moreover, MobileNet V3 (large), which has more parameters, also exhibited higher robustness with an attack success rate of 91.10\% compared to MobileNet V2. However, the attack time and noise rate did not follow a clear pattern as the attack success rate.       

One of the specific improvements that contributed to the increased robustness of MobileNet V3 was the integration of the squeeze-and-excitation (SE) module, which selectively emphasized informative features and suppressed irrelevant ones for more accurate and robust feature extraction. Furthermore, using hard-swish activation functions instead of traditional ReLU activation functions helped mitigate the saturation issue that can occur when using ReLU activations, making the model more resilient to adversarial attacks. 
Lastly, the enhanced batch normalization techniques used in MobileNet V3 helped better normalize the activations within the network, leading to improved performance and robustness.

\section{Discussion}
\label{sec:discussion}

Based on our experiments, we have observed several key findings that can inform the design of deep learning models for improved robustness against adversarial attacks.
Firstly, while batch normalization can improve training speed and stability, it may come at the cost of decreased model robustness. Therefore, when using batch normalization, it is important to carefully evaluate its impact on model robustness and consider alternative normalization techniques such as layer normalization or instance normalization. In \cite{liu2021convolutional}, Liu \etal explored the application of normalization techniques to enhance the training speed and the overall performance robustness of convolutional models. Their results indicated that the models exhibit better performance outcomes by incorporating normalization methods into the training process and demonstrate increased robustness to variations or perturbations in the data. Furthermore, Amini \etal \cite{8970483} suggested that deep learning models' robustness can be enhanced through the implementation of non-smooth regularization techniques.
Secondly, we also noticed that incorporating architectural improvements such as residual connections \cite{he2016deep}, factorization \cite{wang2017factorized}, bottleneck designs \cite{he2016deep}, and squeeze-and-excitation modules \cite{hu2018squeeze} can lead to more robust models. These improvements help the model better capture informative features while suppressing irrelevant ones, leading to better feature extraction and more resilient models. 
Lastly, using activation functions that avoid saturation issues, such as hard-swish instead of traditional ReLU activations, can also improve model robustness.
As a future direction, it would be valuable to evaluate the robustness of updated models across a broader range of architectures, datasets, and attack scenarios \cite{abdukhamidov2022black, abdukhamidov2021advedge,abdukhamidov2022interpretations}. 

\section{Conclusion}
\label{sec:conclusion}

In conclusion, our experiments demonstrate that updates and new versions of deep learning models can have a significant impact on their robustness to adversarial attacks. While batch normalization can improve the training speed and stability of models such as VGG, our results suggest that it may compromise model robustness. On the other hand, updates to the Inception and MobileNet architectures, such as the integration of squeeze-and-excitation modules and improved batch normalization, can lead to improved model robustness. These findings highlight the importance of conducting comprehensive and in-depth research to evaluate the impact of model updates on robustness in critical applications. Future work could explore the generalizability of these results across a broader range of datasets and attack scenarios, and investigate other potential techniques to improve model performance and robustness. 

\section*{Acknowledgment}
This work was supported by the National Research Foundation of Korea(NRF) grant funded by the Korea government(MSIT)(No. 2021R1A2C1011198), (Institute for Information \& communications Technology Planning \& Evaluation) (IITP) grant funded by the Korea government (MSIT) under the ICT Creative Consilience Program (IITP-2021-2020-0-01821), and AI Platform to Fully Adapt and Reflect Privacy-Policy Changes (No. 2022-0-00688).

\balance
\bibliographystyle{IEEEtran}
\bibliography{main}

\begin{thebibliography}{10}
\providecommand{\url}[1]{#1}
\csname url@samestyle\endcsname
\providecommand{\newblock}{\relax}
\providecommand{\bibinfo}[2]{#2}
\providecommand{\BIBentrySTDinterwordspacing}{\spaceskip=0pt\relax}
\providecommand{\BIBentryALTinterwordstretchfactor}{4}
\providecommand{\BIBentryALTinterwordspacing}{\spaceskip=\fontdimen2\font plus
\BIBentryALTinterwordstretchfactor\fontdimen3\font minus \fontdimen4\font\relax}
\providecommand{\BIBforeignlanguage}[2]{{%
\expandafter\ifx\csname l@#1\endcsname\relax
\typeout{** WARNING: IEEEtran.bst: No hyphenation pattern has been}%
\typeout{** loaded for the language `#1'. Using the pattern for}%
\typeout{** the default language instead.}%
\else
\language=\csname l@#1\endcsname
\fi
#2}}
\providecommand{\BIBdecl}{\relax}
\BIBdecl

\bibitem{abuhamad2021large}
M.~Abuhamad, T.~Abuhmed, D.~Mohaisen, and D.~Nyang, ``Large-scale and robust code authorship identification with deep feature learning,'' \emph{ACM Transactions on Privacy and Security (TOPS)}, vol.~24, no.~4, pp. 1--35, 2021.

\bibitem{ali2022effective}
S.~Ali, O.~Abusabha, F.~Ali, M.~Imran, and T.~Abuhmed, ``Effective multitask deep learning for iot malware detection and identification using behavioral traffic analysis,'' \emph{IEEE Transactions on Network and Service Management}, 2022.

\bibitem{chakraborty2021survey}
A.~Chakraborty, M.~Alam, V.~Dey, A.~Chattopadhyay, and D.~Mukhopadhyay, ``A survey on adversarial attacks and defences,'' \emph{CAAI Transactions on Intelligence Technology}, vol.~6, no.~1, pp. 25--45, 2021.

\bibitem{abdukhamidov2023hardening}
E.~Abdukhamidov, M.~Abuhamad, S.~S. Woo, E.~Chan-Tin, and T.~Abuhmed, ``Hardening interpretable deep learning systems: Investigating adversarial threats and defenses,'' \emph{IEEE Transactions on Dependable and Secure Computing}, 2023.

\bibitem{juraev2022depth}
F.~Juraev, E.~Abdukhamidov, M.~Abuhamad, and T.~Abuhmed, ``Depth, breadth, and complexity: Ways to attack and defend deep learning models,'' in \emph{Proceedings of the 2022 ACM on Asia Conference on Computer and Communications Security}, 2022, pp. 1207--1209.

\bibitem{szegedy2014intriguing}
C.~Szegedy, W.~Zaremba, I.~Sutskever, J.~Bruna, D.~Erhan, I.~Goodfellow, and R.~Fergus, ``Intriguing properties of neural networks,'' in \emph{International Conference on Learning Representations, {ICLR} 2014}, Banff, Canada, 2014, pp. 1--10.

\bibitem{barreno2006can}
\BIBentryALTinterwordspacing
M.~Barreno, B.~Nelson, R.~Sears, A.~D. Joseph, and J.~Tygar, ``Can machine learning be secure?'' in \emph{Proceedings of the 2006 ACM Symposium on Information, Computer and Communications Security}.\hskip 1em plus 0.5em minus 0.4em\relax Association for Computing Machinery, 2006, p. 16–25. [Online]. Available: \url{10.1145/1128817.1128824}
\BIBentrySTDinterwordspacing

\bibitem{papernot2017practical}
N.~Papernot, P.~McDaniel, I.~Goodfellow, S.~Jha, Z.~B. Celik, and A.~Swami, ``Practical black-box attacks against machine learning,'' in \emph{Proceedings of the 2017 ACM on Asia conference on computer and communications security}, 2017, pp. 506--519.

\bibitem{goodfellow2015explaining}
I.~J. Goodfellow, J.~Shlens, and C.~Szegedy, ``Explaining and harnessing adversarial examples,'' in \emph{3rd International Conference on Learning Representations, {ICLR} 2015, San Diego, CA, USA}.\hskip 1em plus 0.5em minus 0.4em\relax OpenReview.net, 2015.

\bibitem{kurakin2017adversarial}
A.~Kurakin, I.~Goodfellow, and S.~Bengio, ``Adversarial machine learning at scale,'' in \emph{5th International Conference on Learning Representations, {ICLR} 2017, Toulon, France}.\hskip 1em plus 0.5em minus 0.4em\relax OpenReview.net, 2017.

\bibitem{carlini2017towards}
N.~Carlini and D.~Wagner, ``Towards evaluating the robustness of neural networks,'' in \emph{2017 ieee symposium on security and privacy (sp)}.\hskip 1em plus 0.5em minus 0.4em\relax IEEE, 2017, pp. 39--57.

\bibitem{wang2004ssim}
Z.~Wang, A.~Bovik, H.~Sheikh, and E.~Simoncelli, ``Image quality assessment: from error visibility to structural similarity,'' \emph{IEEE Transactions on Image Processing}, vol.~13, no.~4, pp. 600--612, 2004.

\bibitem{boenisch2021gradient}
F.~Boenisch, P.~Sperl, and K.~B{\"o}ttinger, ``Gradient masking and the underestimated robustness threats of differential privacy in deep learning,'' \emph{arXiv preprint arXiv:2105.07985}, 2021.

\bibitem{szegedy2017inception}
C.~Szegedy, S.~Ioffe, V.~Vanhoucke, and A.~Alemi, ``Inception-v4, inception-resnet and the impact of residual connections on learning,'' in \emph{Proceedings of the AAAI conference on artificial intelligence}, vol.~31, no.~1, 2017.

\bibitem{howard2019searching}
A.~Howard, M.~Sandler, G.~Chu, L.-C. Chen, B.~Chen, M.~Tan, W.~Wang, Y.~Zhu, R.~Pang, V.~Vasudevan \emph{et~al.}, ``Searching for mobilenetv3,'' in \emph{Proceedings of the IEEE/CVF international conference on computer vision}, 2019, pp. 1314--1324.

\bibitem{liu2021convolutional}
S.~Liu, X.~Li, Y.~Zhai, C.~You, Z.~Zhu, C.~Fernandez-Granda, and Q.~Qu, ``Convolutional normalization: Improving deep convolutional network robustness and training,'' \emph{Advances in neural information processing systems}, vol.~34, pp. 28\,919--28\,928, 2021.

\bibitem{8970483}
S.~Amini and S.~Ghaemmaghami, ``Towards improving robustness of deep neural networks to adversarial perturbations,'' \emph{IEEE Transactions on Multimedia}, vol.~22, no.~7, pp. 1889--1903, 2020.

\bibitem{he2016deep}
K.~He, X.~Zhang, S.~Ren, and J.~Sun, ``Deep residual learning for image recognition,'' in \emph{Proceedings of the IEEE conference on computer vision and pattern recognition}, 2016, pp. 770--778.

\bibitem{wang2017factorized}
M.~Wang, B.~Liu, and H.~Foroosh, ``Factorized convolutional neural networks,'' in \emph{Proceedings of the IEEE international conference on computer vision workshops}, 2017, pp. 545--553.

\bibitem{hu2018squeeze}
J.~Hu, L.~Shen, and G.~Sun, ``Squeeze-and-excitation networks,'' in \emph{Proceedings of the IEEE conference on computer vision and pattern recognition}, 2018, pp. 7132--7141.

\bibitem{abdukhamidov2022black}
E.~Abdukhamidov, F.~Juraev, M.~Abuhamad, and T.~Abuhmed, ``Black-box and target-specific attack against interpretable deep learning systems,'' in \emph{Proceedings of the 2022 ACM on Asia Conference on Computer and Communications Security}, 2022, pp. 1216--1218.

\bibitem{abdukhamidov2021advedge}
E.~Abdukhamidov, M.~Abuhamad, F.~Juraev, E.~Chan-Tin, and T.~AbuHmed, ``Advedge: Optimizing adversarial perturbations against interpretable deep learning,'' in \emph{International Conference on Computational Data and Social Networks}, 2021, pp. 93--105.

\bibitem{abdukhamidov2022interpretations}
E.~Abdukhamidov, M.~Abuhamad, S.~S. Woo, E.~Chan-Tin, and T.~Abuhmed, ``Interpretations cannot be trusted: Stealthy and effective adversarial perturbations against interpretable deep learning,'' \emph{arXiv preprint arXiv:2211.15926}, 2022.

\end{thebibliography}

\end{document}